\documentclass[10pt,twocolumn,letterpaper]{article}

\usepackage[pagenumbers]{cvpr} % To force page numbers, e.g. for an arXiv version

\usepackage{bm}
\usepackage{multirow}
\usepackage[table]{xcolor}

\usepackage{algorithm}   
\usepackage{algpseudocode}  % 
\usepackage{amsmath}       %
\usepackage{amssymb}       % 
\usepackage{float}         %
\usepackage{pifont}
\usepackage{fontawesome} %
\usepackage{xcolor}

\definecolor{cvprblue}{rgb}{0.21,0.49,0.74}
\usepackage[pagebackref,breaklinks,colorlinks,allcolors=cvprblue]{hyperref}

\def\paperID{9851} % *** Enter the Paper ID here
\def\confName{CVPR}
\def\confYear{2026}

\usepackage{tikz}
\usepackage{xcolor}

\title{VVS: Accelerating Speculative Decoding for Visual Autoregressive Generation via Partial Verification Skipping}

\author{
Haotian Dong$^{1}$, 
Ye Li$^{1}$, 
Rongwei Lu$^{1}$, 
Chen Tang$^2$,
Shu-Tao Xia$^{1}$,
Zhi Wang$^{1\dagger}$ \\
$^1$Shenzhen International Graduate School, Tsinghua University \\
\quad $^2$The Chinese University of Hong Kong \\
{\tt\small donght24@mails.tsinghua.edu.cn, wangzhi@sz.tsinghua.edu.cn} % Only the first author's email
}

\begin{document}
\maketitle

\begin{NoHyper}
\def\thefootnote{$\dagger$}%
\footnotetext{Corresponding author.}
\end{NoHyper}

\begin{abstract}
Visual autoregressive (AR) generation models have demonstrated strong potential for image generation, yet their next-token-prediction paradigm introduces considerable inference latency.
Although speculative decoding (SD) has been proven effective for accelerating visual AR models, its ``draft one step, then verify one step'' paradigm prevents a direct reduction in the number of forward passes, limiting its acceleration potential.
Motivated by the interchangeability of visual tokens, we explore verification skipping in the SD process for the first time to explicitly cut the number of target model forward passes, thereby reducing inference latency. 
By analyzing the characteristics of the drafting stage, we observe that \textbf{verification redundancy} and \textbf{stale feature reusability} are key factors to maintain generation quality while improving speed for verification-free steps.
Inspired by these two observations, we propose a novel SD framework \textbf{VVS} to accelerate \underline{v}isual AR model via partial \underline{v}erification \underline{s}kipping, which integrates three complementary modules: (1) a verification-free token selector with dynamic truncation, (2) token-level feature caching and reuse, and (3) fine-grained skipped step scheduling.
Consequently, VVS reduces the number of target model forward passes by $\bm{2.8 \times}$ relative to vanilla AR decoding while maintaining competitive generation quality, offering a superior speed–quality trade-off over conventional SD frameworks and revealing strong potential to reshape the SD paradigm.
Our code is available at \url{https://github.com/HyattDD/VVS}.

\end{abstract}
    
\section{Introduction}
\label{sec:intro}

\begin{figure*}[htbp]
  \centering
  \includegraphics[width=0.99\textwidth]{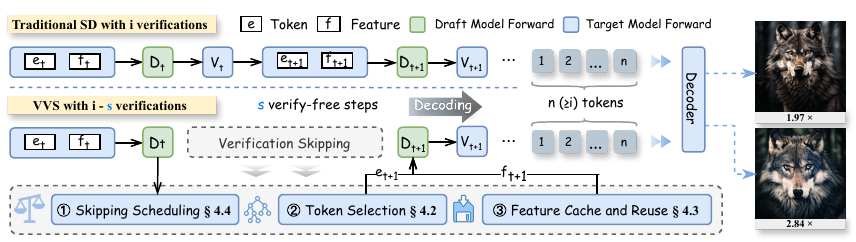}
  \caption{\textbf{Overview of VVS framework}. VVS explicitly reduces the target model forward passes by skipping selected verification stages, thus decreasing the inference latency during SD. 
  Here, $D_t$ and $V_t$ represent the draft and verification stages at iteration $t$, respectively.} 
  \label{fig:overview_VVS}
\end{figure*}

Large language models (LLMs) based on autoregressive (AR) architecture \cite{yang2025qwen3technicalreport,deepseekai2025deepseekv3technicalreport,dong2025survey} have shown exceptional performance in text generation and demonstrated significant potential in visual generation, showing competitive performance against the diffusion models~\cite{DeDiff2020,yang2023diffusion,luo2025dice}, offering promising prospects for developing unified multi-modal models~\cite{liu2024luminamgpt,sun2024llamagen,xin2025luminamgpt2,emu3,chern2024anole,chameleon2025chameleon}.
However, AR generation is inherently sequential, requiring hundreds or thousands of forward passes. This leads to higher latency in interactive image generation, which requires full token processing, unlike streaming-based text generation.

To accelerate the decoding of AR models, numerous techniques have been investigated~\cite{chen2025collaborative,fu2024break,del2023skipdecode,stern2018blockwise,he2025zipar}; among them, speculative decoding (SD) has emerged as a promising and widely adopted approach~\cite{zheng2025faster,cai2024medusa,li2024eagle2,miao2024specinfer}.
SD utilizes a compact draft model to efficiently generate multiple candidate tokens within a significantly short time frame.
The target model (the AR model to be accelerated) then verifies these tokens in parallel, accepting multiple tokens per forward pass.
During verification, multiple tokens can be accepted within a single forward pass.
Even if all candidate tokens are rejected, the target model can still obtain at least one token by sampling, with only minimal overhead from waiting for the draft model to generate candidates.

Since language and visual tokens follow markedly different semantic distributions, naïvely applying SD to visual AR models typically yields a low acceptance rate for the draft model’s candidate tokens.
Based on the interchangeability of visual tokens, existing work leverages clustering~\cite{so2025GSD} and relaxation~\cite{jang2024lantern} to avoid excessive rejection caused by strict verification, thereby achieving a high token acceptance ratio and speedup for the visual AR model.
However, they still conservatively verify every proposal from the draft model, so \textit{the calling times of the target model are not explicitly reduced, limiting inference speed}.

In this paper, rather than directly adopting the traditional SD paradigm from text to vision, we take a step further by introducing partial verification skipping for drafted candidates.
However, skipping verification during SD is not a free lunch; we face a series of issues: 
\ding{172} How to select the accepted tokens without verification? 
\ding{173} How to proceed to the next round of drafting? Skipping verification means that no intermediate states are generated in the current SD step.
\ding{174} How should we distribute the verification-free steps? Excessive bypassing of verification turns the draft model into the primary generator, inevitably causing severe quality degradation.
To tackle these challenges, we propose VVS, a novel SD framework tailored for visual AR generation models, which enables partial skipping of verification steps during the SD process.

Two key observations underpin our framework design: \textbf{(i)} verification redundancy and \textbf{(ii)} stale feature reusability.
During the SD drafting stage, the draft model proposes several candidate tokens for every position, building a candidate-token tree that contains multiple paths to be verified. 
We observe that these paths exhibit highly visual similarity during most iterations (\textit{e.g.}, in 75\% of SD iterations, sequences show a cosine similarity exceeding 0.7). 
Such prevalence of similarity implies that exhaustive verification of all paths may not consistently yield visually distinct outputs, thereby revealing a degree of verification redundancy.
Moreover, throughout the entire token-sequence generation process, the features employed for drafting in consecutive SD iterations remain highly similar (\textit{e.g.}, the similarity between features of adjunct tokens is 0.68), so reusing slightly stale features still yields dependable candidate token sequences.
The first observation demonstrates that verification-free token selection is viable, confirming the feasibility of bypassing part verification steps. 
The second one shows that the draft model can still deliver reasonably acceptable candidates even without fresh intermediate states from the target model.
These two observations are critical to preserving generation quality and speedup when verification steps are omitted during SD.

Building upon these two key observations, we propose a novel SD framework VVS to accelerate visual AR generation via partial verification skipping, as shown in \cref{fig:overview_VVS}.
VVS integrates three complementary modules: 
(1) a verification-free token selector to choose candidate tokens to be accepted, 
(2) a stale feature caching and reuse mechanism to compensate for the missing intermediate states for drafting, and 
(3) a fine-grained skipping-step scheduler to determine verification-free steps based on the token path similarity of the draft token tree.
As a result, VVS achieves an average of $\bm{2.8\times}$ target model forward passes reduction over vanilla decoding with minimal degradation of generation quality and achieves better speedup-quality trade-off compared to traditional SD frameworks, showing significant potential to revolutionize the traditional SD paradigm.

Our contributions are summarized as follows: 
\begin{itemize}
    \item We are the first to explore partial verification skipping during SD and we propose a novel SD framework tailored for visual AR generation to reduce the target model forward passes explicitly.
    \item By analyzing token path similarity during candidate token tree construction and feature similarity throughout the generation process, we reveal \textit{verification redundancy} and \textit{stale feature reusability} in the drafting stage.
    \item Inspired by our analysis, we introduce VVS, a novel SD framework tailored for visual AR generation models. VVS dynamically determines verification-free steps, directly accepts selected tokens, and reuses cached stale features to propose candidate continuations, thereby bypassing a portion of the verification stages.
    \item Our framework VVS yields $\bm{2.8 \times}$ fewer target model forward passes over vanilla AR decoding with minimal degradation in generation quality and trades off better between quality and efficiency.
\end{itemize}

\section{Preliminary}
\label{sec:prel}

\paragraph{Visual AR Generation Model} leverages a sequential token prediction scheme~\cite{brown2020language} to synthesize high-fidelity visual content. 
During inference, visual AR generation models predict subsequent tokens $ \bm{x_{s+\ell}} $ conditioned on prior tokens $ \bm{x_{1:s+\ell-1}} $, adhering to the probability distribution $ P(\bm{x_{s+1:s+k}} | \bm{x_{1:s}}) = \prod_{\ell=1}^{k} P(\bm{x_{s+\ell}} | \bm{x_{1:s+\ell-1}}) $, where $ k $ denotes the total number of image tokens. The resulting token sequence is subsequently processed by a pre-trained visual decoder (\textit{e.g.}, VQ-VAE~\cite{oord2018vqvae}, VQ-GAN~\cite{yu2022vqGAN}) to reconstruct the final RGB image, enabling versatile generation across vision-language tasks.

\paragraph{Speculative Decoding} \citep{leviathan2022fast,chen2023accelerating} initially proposed SD to accelerate the inference process of AR language models.
SD leverages a smaller draft model $M_d$ to generate candidate tokens $ \{x_{s+1}, x_{s+2}, \dots, x_{s+\gamma-1}\} $, in which each token is sampled from distribution $p(x_{s+i})$. Then these candidates are concatenated with the input sequence $x_s$ and passed to the larger target model $M_t$ to obtain the probabilities $\{q(x_{s+1}), q(x_{s+2}), \dots, q(x_{s+\gamma})\}$ in parallel.
During sequential verification, the acceptance probability for candidate token $x_{s+i}$ follows $\text{min} \{1, {q(x_{s+i})}/{p(x_{s+i})}\}$. Once a token is rejected, it will be resampled from the adjusted distribution $\text{max} \{0,p(x_{s+i})-q(x_{s+i})\}$. If all candidates are accepted, one more token will be sampled from $q(x_{s+\gamma})$.
We use $q(x_{s+i})$ and $p(x_{s+i})$ to denote $q(x_{s+i}|\bm{x_{<i+s}})$ and $p(x_{s+i}|\bm{x_{<i+s}})$ for simplicity.
To accelerate visual AR models with SD, \citep{jang2024lantern} relaxes the acceptance probability for a candidate token \(x_{s+i}\) to \(\min\!\left\{1,\; \sum\nolimits_{i=1}^{k} q(x_{s+i})\,/\,p(x_{s+i})\right\}\), where \(\sum\nolimits_{i=1}^{k} q(x_{s+i})\le \delta\) and \(\delta\) is the relaxation threshold that pools the probabilities of the top-k neighboring tokens.

\begin{figure}[tbp]
  \centering
  \begin{subfigure}[b]{0.235\textwidth}
    \centering
    \includegraphics[width=\linewidth]{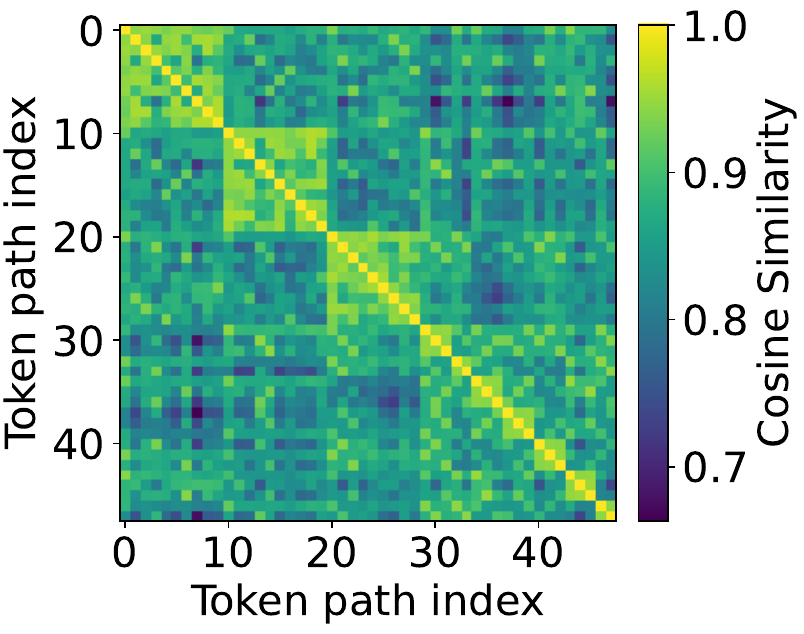}
    \caption{}
    \label{fig:minipage1}
  \end{subfigure}%
  \hfill
  \begin{subfigure}[b]{0.235\textwidth}
    \centering
    \includegraphics[width=0.87\linewidth]{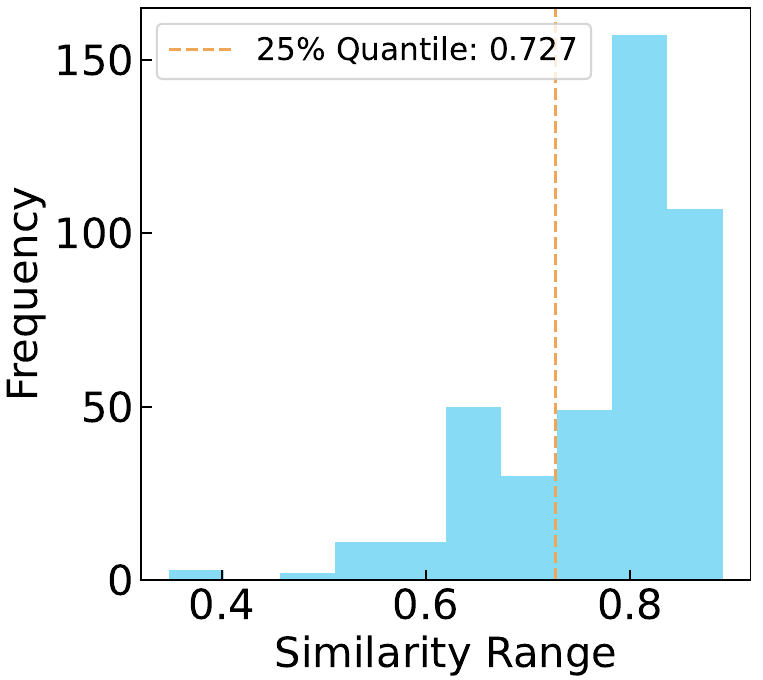}
    \caption{}
    \label{fig:minipage2}
  \end{subfigure}
  \caption{Similarity of the drafted candidate token tree. (a) Visual similarity among different token paths within a candidate token tree. (b) Average similarity distribution of token trees.}
  \label{fig:token_path_sim}
\end{figure}

\section{Drafting Stage Analysis}
\label{sec:analysis}

In the traditional SD paradigm, the draft model interacts with the target model in two ways: it first forwards the previously verified tokens and hidden state so the draft model can generate candidate continuations, and then submits those candidates for verification.
We analyze the dynamic characteristics of the drafting stage during SD in visual AR generation, exploring whether candidates and their acceptance can occur without the target model involving in.

\paragraph{Verification Redundancy.} 
\label{para:observe_token}
Tree-based SD enables the large target model to simultaneously verify multiple candidate tokens at each position proposed by the draft model, accepting the token path that passes validation.
Prior work observed that the draft model assigns low confidence to visual tokens, which are largely interchangeable~\cite{jang2024lantern}; accordingly, a position-wise relaxed acceptance mechanism was introduced, while still performing full verification of token paths.
Going beyond individual positions, we further analyzed the similarity among the various token paths in the candidate tree. Specifically, for each path, we computed exponentially decayed weighted cosine similarities across token positions. 
Across most SD steps, the paths in the draft token tree display high mutual similarity, as shown in \cref{fig:token_path_sim}.
This similarity implies that the draft model provides the target model with visually similar token paths, thereby weakening the gains through exhaust verification.

\begin{figure}[t]
    \centering
    \subfloat[Feature average similarity vs. feature Staleness.]
    {%
        \includegraphics[width=0.98\columnwidth]{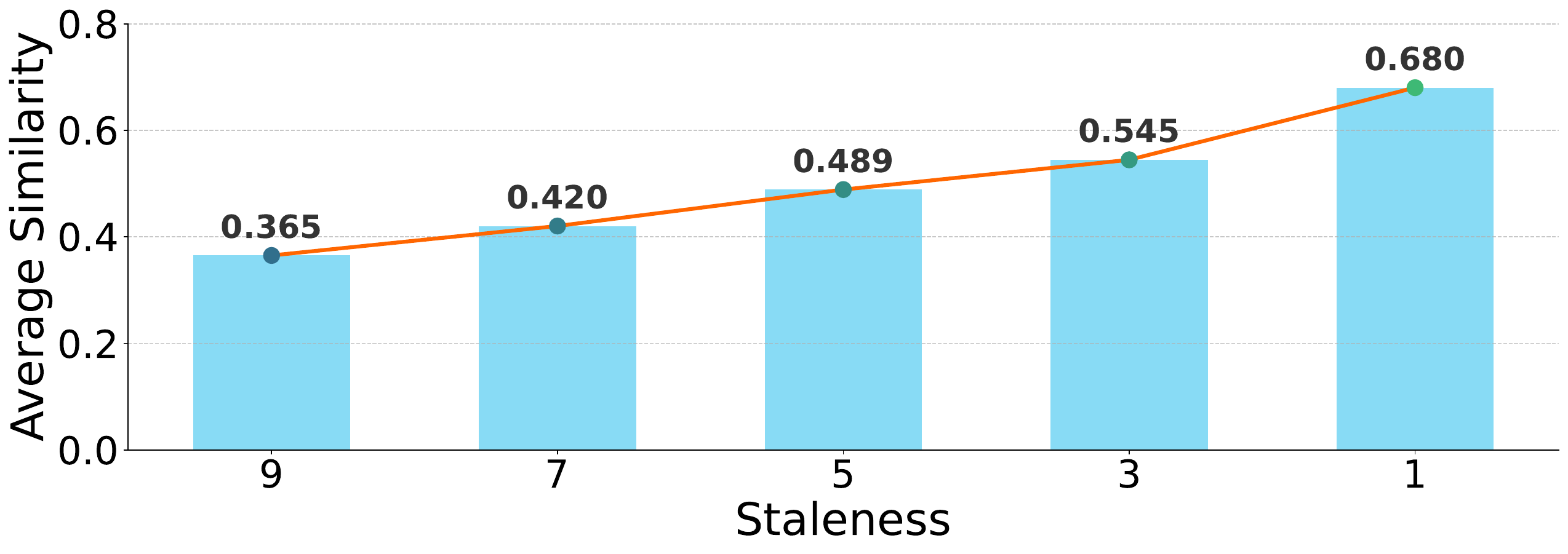}%
        \label{fig:feat_sim_trend}%
    }
    \vfill 
    \subfloat[Token-wise feature similarity during generation.]
    {%
        \includegraphics[width=0.98\columnwidth]{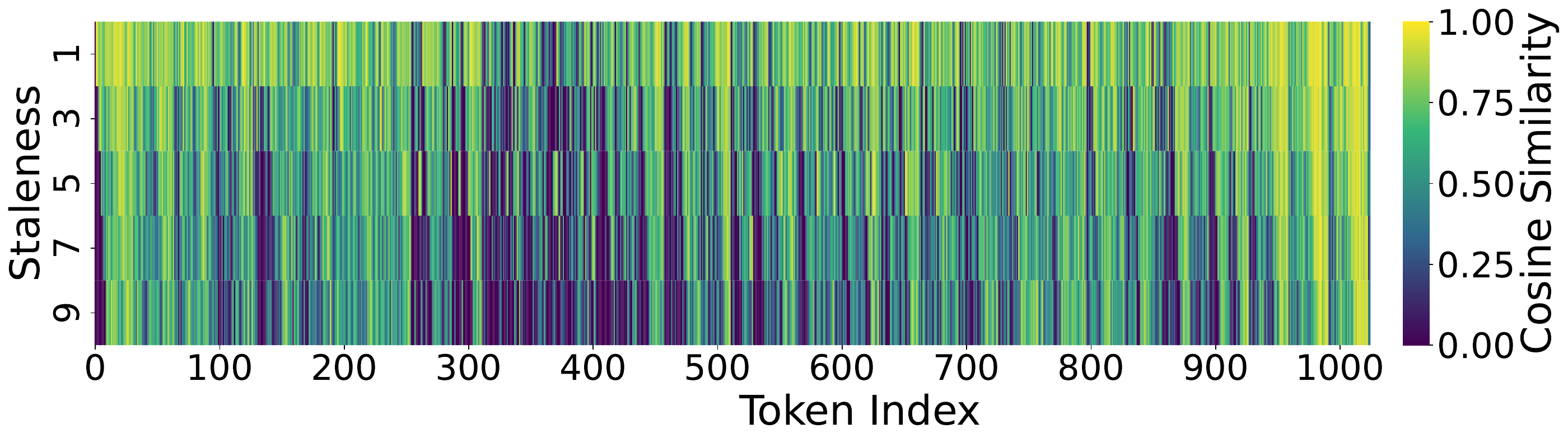}%
        \label{fig:feat_sim_heatmap}%
    }

    \caption{Token-level feature similarity across different staleness during SD generation. As staleness increases, the similarity between stale features and fresh features decreases progressively.}
    \label{fig:similarity_analysis}
\end{figure}

To test this conjecture, we conducted experiments in which we fixed the verification result to a specific token path while preserving the relaxed verification mechanism.
We find that the visual AR model generation quality is insensitive to deliberate perturbations of the verification result: dynamically substituting the token path selected by the target model with an alternative one rarely harms image fidelity.
This observation implies that, \textit{during SD of visual AR generation, exhaustive verification for all steps is not strictly required to maintain quality and efficiency---verification contains a certain degree of redundancy.}

We further perform a finer-grained investigation by controlling the proportion $r$ of target model verification results that are replaced during the entire image generation process. 
Verification result interchangeability yields robustness across $r$ from 0 \% to 100 \%, and the quality of image generation and the acceleration of SD can remain stable (\textit{e.g.}, FID hovers within a 0.4 width band under the same speedup, more details are provided in the supplementary material).
It should be noted, however, that relying solely on the draft model for generation is not reliable, since the features generated from the verification of the target model are still indispensable for producing acceptable candidates.

\begin{figure}[tbp]
  \centering
  \includegraphics[width=0.47\textwidth]{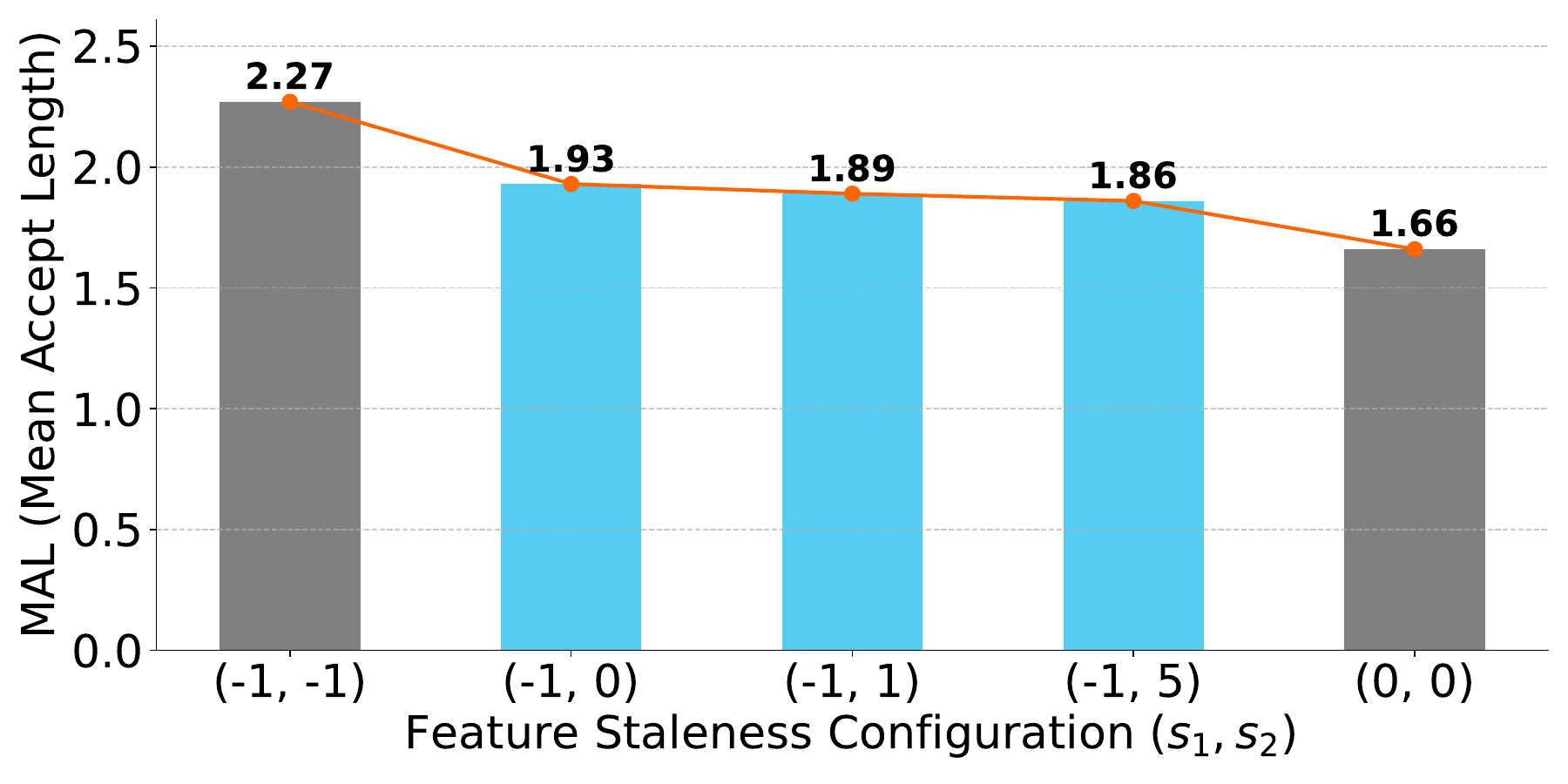}
  \caption{Mean accept length comparison under feature blending with different additional staleness. $s$ denotes the extra staleness introduced for cached features; $s=-1$ indicates using the freshest features; $s=0$ indicates using the most recent features cached from prior steps; $s=i~(i>0)$ indicates using features with additional staleness $i$ compared with $s=0$.}
  \label{fig:feature_blend}
\end{figure}

\paragraph{Stale Feature Reusability.}
\label{para:observe_feature}
Leveraging the features (hidden states before output head) generated during the verification process of target model to guide tree-based candidates generation yields remarkable acceptance rates in language modeling \cite{cai2024medusa,li2024eagle1,li2025eagle3}. 
Specifically, the draft model utilizes the accepted tokens and their features verified at step $i-1$ to draft candidate tokens at step $i$.

We first examined how token-level features evolve during SD. As Figure 3 illustrates, feature similarity declines with increasing token distance, yet neighboring tokens still achieve a similarity of 0.68, indicating that stale features remain partly reusable. 
To verify this, we adopt SD with the feature-based drafting paradigm \cite{li2024eagle2} for visual AR models and find that the draft model can still provide acceptable candidates with stale features.

Specifically, we cache the features of each token during the verification stage; when proposing candidates at step~$i$, the draft model reuses the last $n$ features cached before step~$i-1$  with the corresponding $n$ tokens accepted at step $i-1$.
Then we find that stale features before step $i-1$ still enable the draft model to propose acceptable tokens for step $i$.
Although stale features at the last step ($s=0$) introduce some degradation, the mean accept length (MAL) still remains at 73\% relative to fresh features (more details in the supplementary material).
Crucially, we observe that features leveraged during SD can be blended across varying staleness levels, as shown in~\cref{fig:feature_blend}.
We alternately utilize features from two distinct staleness categories, each constituting 50\% of the generation process.
As a result, blending fresh features with stale ones raises the MAL maintainability from 73\% to 85\% compared with purely using fresh features.
This suggests that \textit{stale features produced during SD remain reusable, delivering comparable acceleration}.

\section{Methodology}
\label{sec:method}

\begin{figure*}[t]
\centering
\includegraphics[width=0.99\textwidth]{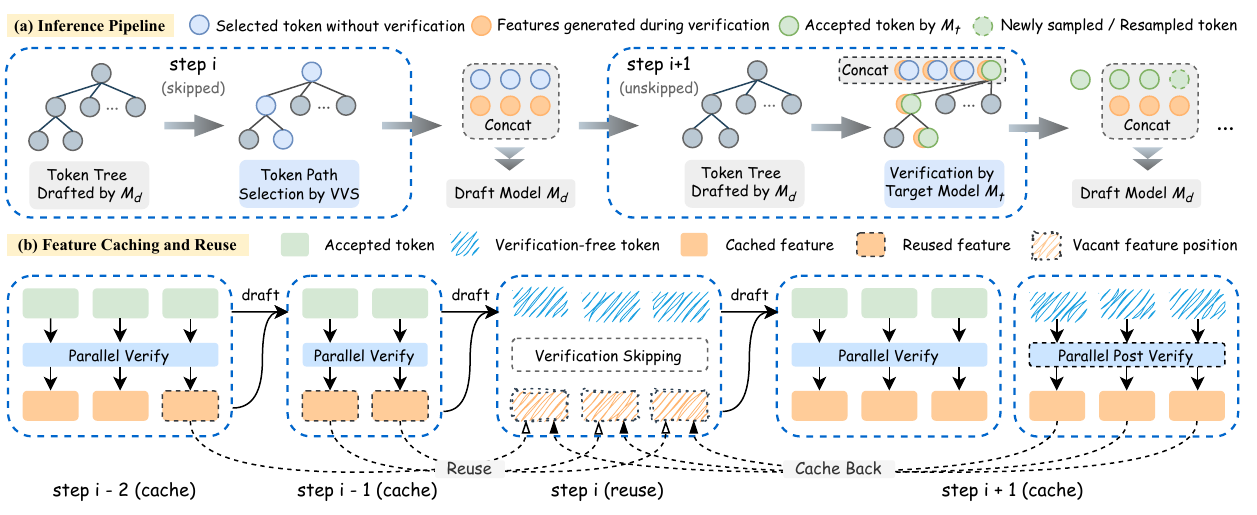}
\caption{\textbf{(a) Inference pipeline of our SD framework VVS}, which supports partial verification skipping. \textbf{(b) Token-level feature caching and reuse mechanism.} Since the number of tokens accepted at different iterations varies and truncation in~\cref{sec:4.2} is applied, the cached features to be reused could come from multiple steps. Tokens accepted without verification proceed to the target model at the next verification step—which we term \textit{post verification}—and their resulting features are cached back to the corresponding positions.}
\label{fig:vvs_pipeline}
\end{figure*}

Inspired by the observations in \cref{sec:analysis}, we propose \textbf{VVS}: a novel SD framework for visual AR generation that bypasses the verification partially, explicitly cutting target model forward passes to achieve extra acceleration.
We devised a candidate token selector when verification is skipped, paired with a feature caching and reuse scheme to keep the drafting process on track.
To enable a more fine-grained trade-off between generation quality and speedup, we also designed a scheduler to allocate skipping steps dynamically.

\subsection{Partial Verification-Skipping Pipeline}
\label{sec:4.1}
 
When verification skipping is activated at step~$i$, the draft model produces candidate tokens $\bm{c_i}$ conditional on:
(i)~the stale features $\bm{h}_{i-1}$ cached from prior steps and
(ii)~the embedding $\bm{e}_{i-1}$ of selected token sequence $\bm{x}_{i-1}^{*}$.
Then one token sequence of $\bm{c_i}$ is accepted immediately without querying the target model, yielding the tentative continuation~$\bm{x}_i^{\circ}$.
At the subsequent verification-resumed step~$i+1$, the unverified sequence~$\bm{x}_i^{\circ}$ is prepended to the newly drafted candidates $\bm{c}_{i+1}$, then the concatenated sequence is processed in a single forward pass of the target model.
This process, which we term \textit{post verification}, simultaneously processes old unverified tokens alongside new ones. By restoring the missing KV-cache entries, it reestablishes exact AR conditioning.
Besides, features ~$\bm{h}_t$ of any step $t$ are cached in a buffer, ensuring that the draft model always has access to the most recent verified representation whenever verification skipping is enabled.
\cref{fig:vvs_pipeline} depicts the pipeline of VVS, and we provide pseudocode \cref{alg:skip_speculative} in the supplementary material.

A noticeable condition is that if verification skipping is used consecutively, $\bm{x}_{i-1}^{*}$ could be $\bm{x}_{i-1}^{\circ}$. 
This can result in too many unverified tokens, thereby causing uncontrollable error accumulation.
To bound error accumulation, we enforce a hard constraint that no two consecutive steps can be verification-free.

\subsection{Unverified Token Selection with Truncation}
\label{sec:4.2}

To ensure quality during verification-free steps—where no alignment with the target model occurs—we must safeguard the tokens we accept. 
The simplest strategy is to trust the draft model outright and follow the candidate path whose confidence is highest.
Yet the draft model constructs its token tree greedily; relentlessly picking the most confident path injects greedy-decoding fragments into the overall process, visibly degrading image fidelity.
To match the diversity of decoding under temperature=1, we abandon deterministic selection and instead sample uniformly from the candidate paths, converting every verification-free step into a stochastic draw.
To demonstrate the efficacy, we ablate different token path selection strategies in~\cref{sec:ablation}.

Since cumulative confidence decays along the token path, preserving every token in a long trajectory erodes visual fidelity, and the token tree structures produced by dynamic pruning steps differ~\cite{li2024eagle2}.
We therefore dynamically truncate the selected token path to its first
\begin{equation}
\gamma = \min\Bigl(L_s,\, \bigl\lfloor \bar{L} \bigr\rfloor \Bigr),
\qquad
\bar{L} = \frac{1}{|\mathcal{P}|} \sum_{P_i \in \mathcal{P}} |P_i|,
\end{equation}
\label{eq:trunc}
where $L_s$ is the length of the selected token path, $\bar{L}$ is the average length of the paths in the pruned token tree, and $\mathcal{P}$ denotes the set of these paths.
\Cref{fig:acp_len} shows that the token acceptance length evolves differently before and after applying the truncation strategy. As the draft model’s confidence in successive candidate tokens declines, limiting the number of skipped tokens helps maintain image quality.

\subsection{Token-level Feature Caching and Reuse}

Enabling verification skipping deprives the next drafting phase of the features needed to construct the candidate tree.  
Leveraging the observation that stale features are reusable from~\cref{sec:analysis}, we cache token-level features produced during each verification pass and reuse them whenever drafting proceeds without verification of the last step.
Since the number of accepted tokens $\bm\gamma_i$ fluctuates across steps, the retrieved features may span multiple prior steps rather than the single latest verified step, as illustrated in \cref{fig:vvs_pipeline}.  
We retrieve from the cached features $\bm{f}_{i-1}$ to obtain the exact number of most recent token-level features (without additional staleness) that matches the number of tokens accepted without verification at the $i$-th step, denoted as $\bm\gamma_i$. 
Consequently, the entire SD process with partial verification skipping operates with features of mixed staleness.
This heterogeneous reuse scheme underpins the partial verification-skipping SD pipeline, and the empirical validity of stale feature reusability substantiates its practicality.
We conducted an ablation study on feature staleness in ~\cref{sec:ablation}.

\subsection{Fine-grained Skipped Step Scheduling}

Since skipping verification affects both the accepted tokens and the quality of subsequent draft steps, controlling when and how often verification is skipped is essential for balancing generation quality and speed. 
We explore two scheduling policies for skipped steps. The first strategy uses a fixed interval $i$: after $i-1$ normal verified steps, the $i$-th verification step is skipped. This rule-based approach introduces almost no runtime overhead, yet it ignores the dynamics of the candidate token tree at each step and offers only coarse-grained control over the skip frequency. 
Motivated by the observations in \cref{para:observe_token}, we propose a dynamic strategy that skips verification when the similarity among different token paths in the candidate tree exceeds a threshold $\bar{S}$, defined as
\begin{equation}
\bar{S}=\sum_{\ell=0}^{L-1} w_{\ell}\cdot S^{(\ell
)}, \quad
\text{with} \quad
w_{\ell}=\frac{\alpha^{\ell}}{\sum_{k=0}^{L-1}\alpha^{k}}.
\end{equation}
\label{eq:sim_compute}

where $\alpha$ is an exponential decay weight that controls the contribution of the cosine similarity of token representations at each position $\ell$.
The rationale is that highly similar candidates are likely to yield a token close to the verified one even if verification is bypassed. 

However, computing this similarity at every iteration incurs noticeable decision overhead. 
We therefore accelerate the computation with \texttt{torch.jit} and token path down-sampling, resulting in only around 25\% time cost than before, as detailed in \cref{alg:dynamic_skip}. 
To prevent error accumulation, we further enforce a minimum separation of one verified step between consecutive skips.

\renewcommand{\algorithmiccomment}[1]{\hfill \(\triangleright\) #1}

\begin{figure}[tbp]
\begin{algorithm}[H]
\small
\caption{Dynamic Verification Skipping Strategy}
\label{alg:dynamic_skip}
\begin{algorithmic}[1]
\Require
    $\mathcal{C} = \{c_1, ..., c_N\}$: candidate token sequences of current iteration;
    $\mathcal{E}$: latent embedding codebook;
    $T_{\text{sim}}$: similarity threshold;
    $\bar{S}$: average similarity of candidate token paths;
    $V_{\text{last}}$: boolean, whether the last step was verified;
\Ensure
    $\text{decision} \in \{\text{True}, \text{False}\}$: whether to bypass verification

\If{$\neg\ V_{\text{last}}$}                      \Comment{Once skip, verify next step}
    \State \Return \text{False}
\EndIf

\State $\mathcal{C}_{\text{sub}} \leftarrow \text{Downsample}(\mathcal{C}, \text{stride}=2)$
\Comment{Downsample}

\State $\bar{S} \leftarrow \text{CompSim} 
(\mathcal{C}_{\text{sub}}, \mathcal{E})$
% \texttt{@torch.jit}  
\Comment{\texttt{@torch.jit}}

\If{$\bar{S} < T_{\text{sim}}$}     \Comment{Low similarity, do verify}
    \State \Return \text{False}             
\Else
    \State \Return \text{True}              \Comment{High similarity, skip verification}
\EndIf
\end{algorithmic}
\end{algorithm}
\end{figure}

\section{Experiment}
\label{sec:exp}

In this section, we demonstrate the effectiveness of the proposed SD framework VVS, showcasing its ability to accelerate the SD process for visual AR models.
We evaluate the proposed VVS framework on two visual AR generation models: LlamaGen-XL Stage I and Stage II~\cite{sun2024llamagen}.
We generate images from the MS-COCO validation captions~\cite{Lin2014MicrosoftCOCO} and benchmark their quality against the corresponding ground-truth images. 
To measure quality and speedup, we leverage all 5,000 samples from the MS-COCO evaluation dataset.
Unless otherwise specified, experiments are conducted on a server equipped with $8\times$ RTX 4090 GPU and $2\times$ Intel Xeon Platinum 8563C CPUs.

\paragraph{Baselines}

We compare VVS against SD baselines integrated with EAGLE-2~\cite{li2024eagle2} that do not support partial verification skipping.
Since vanilla SD with strict verification fails to accelerate visual AR models~\cite{jang2024lantern,so2025GSD}, we incorporated the relaxed acceptance mechanism~\cite{jang2024lantern} into VVS.
A vanilla AR decoding baseline is also included for reference.
We fix the decoding temperature at 1.0 and omit greedy decoding (temperature = 0), since it markedly degrades the visual fidelity of AR-generated images~\cite{teng2024SJD,so2025GSD}, unlike its effect in LLMs.
To ensure a fair comparison, all SD baselines use the same drafter checkpoints from~\cite{jang2024lantern}.

\paragraph{Metrics}
{\textbf{(i) Speedup Metrics.}} 
    We measure speedup by the average number of \underline{t}okens produced \underline{p}er target model \underline{f}orward pass,
    \begin{equation}
    \mathrm{TPF}=
    \frac{\text{\#Generated tokens}}{\text{\#Target model forward passes}},
    \end{equation}
    and quantify end-to-end inference speedup~$l$ by wall-clock time.
    $\mathrm{TPF}$ is independent of the computing device, whereas $l$ is tied to it.
    For baselines that do not support partial verification skipping, TPF essentially aligns with MAL, which refers to the average number of tokens accepted per iteration.
{\textbf{(ii) Quality Metrics.}}  
    VVS trades off generation quality against speed. We quantify the generation quality with FID~\cite{heuselGAN2017_fid}, CLIP score~\cite{hessel-etal-2021-clipscore}, Precision and Recall~\cite{Kynknniemi2019precision_recall}, and HPS v2~\cite{wu2023hpsv2} during tasks above.

\begin{figure*}[t]
\centering
\includegraphics[width=0.99\textwidth]{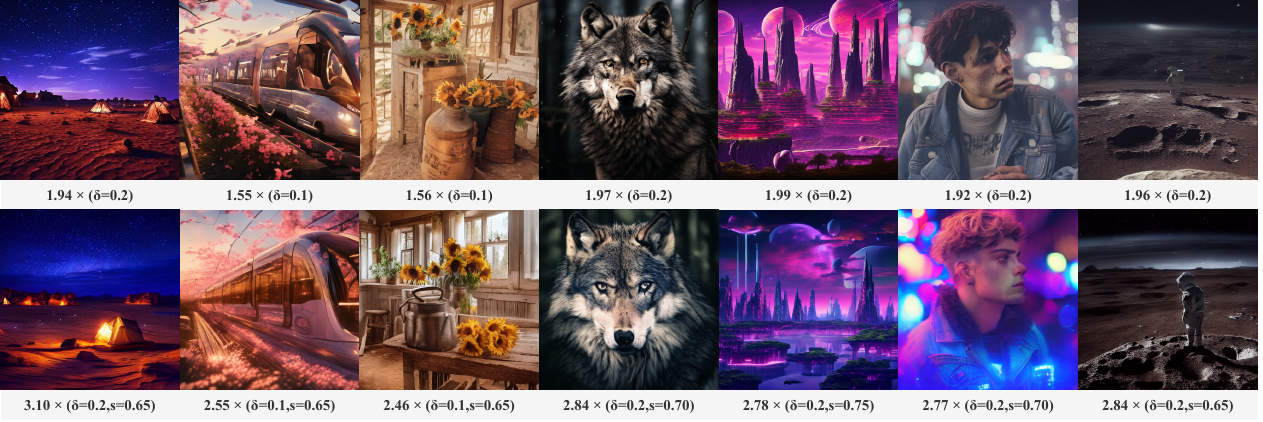}
\caption{Qualitative comparison of results between the acceptance-relax-based SD framework (upper) and ours (lower). $\delta$ denotes the probability threshold for relaxed acceptance, while $s$ is the candidate token path similarity threshold used to determine whether the verification step can be skipped. The prompts used to generate these images are provided in the supplementary material.}
\label{fig:example_figures}
\end{figure*}

\subsection{Main Results}

\paragraph{Quantitative Results}

To evaluate the efficacy of VVS in accelerating visual AR generation speed while maintaining quality, we conducted experiments on different baselines, as shown in \cref{tab:main_results}. 
VVS cuts the target model’s forward passes by up to $\bm{2.86\times}$ and delivers up to $\bm{1.76\times}$ wall-clock inference speed-up, outperforming the baseline in speed while needing only a slight acceptance relaxation of $\delta$ = 0.1, while preserving or even enhancing generation quality.
These experimental results show that VVS, as a new SD paradigm, can accelerate visual AR models by explicitly reducing the target model forward passes.

\paragraph{Qualitative Results}

To demonstrate that VVS accelerates visual AR generation without compromising quality, we produced a variety of prompts from diverse styles, including landscape, portrait, animal, and both indoor and outdoor scenes, as illustrated in \cref{fig:example_figures}.
With dynamic verification skipping (under similarity threshold $s=0.65$), VVS can achieve up to $\bm{3.1\times}$ TPF under the relaxation threshold of~$\delta=0.2$ without visual degradation.
This confirms that our partial verification-skipping paradigm unlocks a new acceleration route for SD on visual AR models, surpassing mere relaxed acceptance.

\paragraph{Trade-off Between Speed and Quality}

\begin{figure}[tbp]
  \centering
  \hspace*{-0.3cm}
  \begin{minipage}{0.35\textwidth}
  \includegraphics[width=\linewidth]{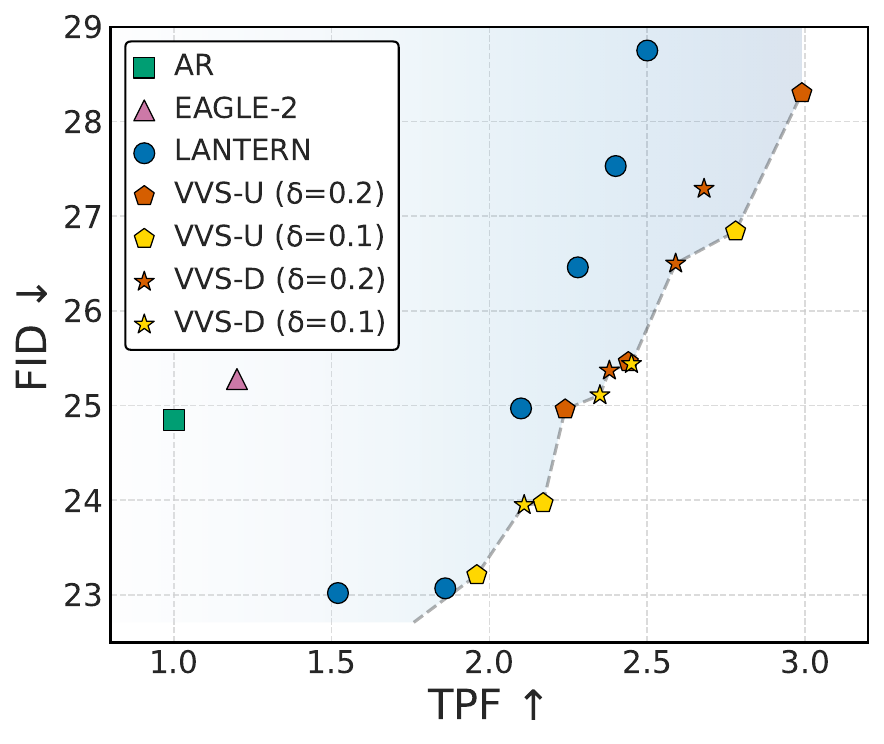} 
  \end{minipage}
    \caption{Pareto-front Comparison: TPF vs FID. AR denotes vanilla autoregressive decoding. VVS-U denotes skipping verification uniformly, VVS-D denotes skipping verification dynamically based on candidate token path similarity. $\delta$ is the relaxed probability threshold during verification.}
  \label{fig:pareto}
\end{figure}

We conduct supplementary experiments to demonstrate that VVS can achieve a better trade-off between quality and speed, as shown in \cref{fig:pareto}. For LANTERN, we vary the relaxation threshold $\delta$ from 0.1 to 0.6 to obtain different acceleration rates. In the VVS experiments, we fix the relaxed acceptance threshold $\delta$ at 0.1 and 0.2, and adjust the number of skipped verification steps for both uniform skipping and dynamic skipping strategy to control the speedup. 
For uniform skipping (VVS-U), the skipping interval $i$ is set to 4, 3, and 2. For similarity-based dynamic skipping (VVS-D), the similarity threshold $\bar{S}$ on candidate token paths is set to 0.70, 0.75, and 0.80. 
The results show that VVS yields a superior FID–TPF trade-off, and the similarity-based schedule enables finer-grained control, compensating for the quality instability inherent in uniform skipping.

\subsection{Ablation Study}
\label{sec:ablation}

\paragraph{Token Selection and Truncation}

\begin{figure}
    \centering
    \includegraphics[width=0.98\linewidth]{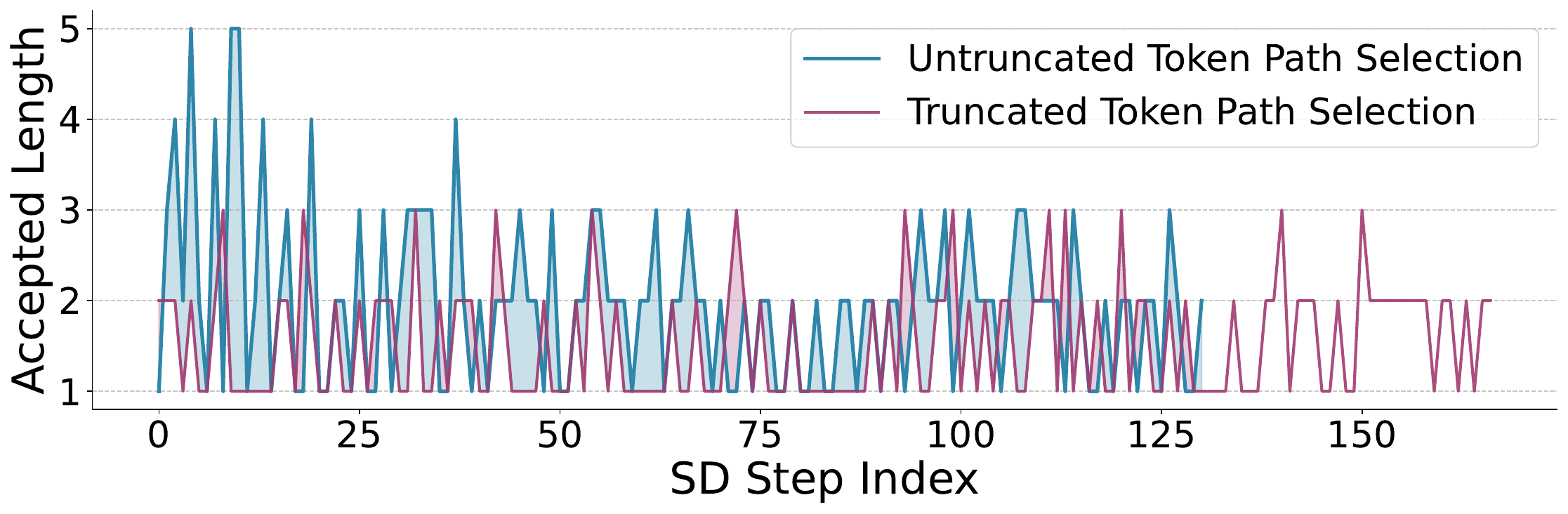}
    \caption{Accept length dynamically changes during the SD process. With token path dynamic truncation under verification skipping, the accept length cross steps are more stable.}
    \label{fig:acp_len}
\end{figure}

When verification is skipped, we compare two token selection strategies: (i) directly selecting the token path with the highest confidence, (ii) uniformly sampling one token path.
Then we truncate the token sequence according to the average length of all candidate paths. 
In \cref{tab:abla_token} we compare the two token-path selection strategies. 
Both yield comparable TPF acceleration, yet uniform sampling better preserves image fidelity. 
We attribute this to the greedy construction of the token tree: a verification-free step with fixed selection commits to the greedy continuation, whereas uniform sampling introduces stochasticity that injects beneficial diversity.
We also conducted an ablation study on the truncation mechanism. 
Without truncation, the number of accepted tokens fluctuates widely (\cref{fig:acp_len}), leading to excessive tokens bypassing verification. 
While this yields higher acceleration, it significantly degrades image quality.

\begin{table*}[ht]
\centering
\caption{Quantitative evaluation on MS-COCO 2017. Wall-clock latency speedup $l$ is measured on a single NVIDIA A40 GPU.
}
\label{tab:main_results}
\setlength{\tabcolsep}{6pt}
\resizebox{490pt}{!}
{

\begin{tabular}{lcccccccc}

\toprule
\multirow{2}{*}{\textbf{Model}} &
\multirow{2}{*}{\textbf{Method}} &
\multicolumn{2}{c}{\textbf{Speedup Metrics} ($\uparrow$)} &
\multicolumn{5}{c}{\textbf{Image Quality Metrics}} \\

\cmidrule(lr){3-4} \cmidrule(lr){5-9}
 & & \textbf{$\boldsymbol{l}$} & \textbf{TPF} &
\textbf{FID ($\downarrow$)} &
\textbf{CLIP score ($\uparrow$)} &
\textbf{Precision ($\uparrow$)} &
\textbf{Recall ($\uparrow$)} &
\textbf{HPSv2 ($\uparrow$)} \\
\midrule

\multirow{4}{*}{Llamagen-XL Stage I} 
& Vanilla AR & 1.00$\times$ & 1.00 & 24.88 & 0.3220 & 0.5084 & 0.6684 & 0.2350 \\
& EAGLE-2~\cite{li2024eagle2} & 0.87$\times$ & 1.22 & 25.28 & 0.3220 & 0.5094 & 0.6690 & 0.2353 \\
& LANTERN ($\delta=0.3$)~\cite{jang2024lantern} & 1.45$\times$ & 2.10 & 24.97 & 0.3223 & 0.5494 & 0.6438 & 0.2293 \\
& VVS-U ($i=4, \delta=0.2$) & \textbf{1.63$\times$} & \textbf{2.24} & 24.96 & 0.3210 & 0.5416 & 0.6486 & 0.2241 \\

\midrule

\multirow{4}{*}{Llamagen-XL Stage II}
& Vanilla AR & 1.00$\times$  & 1.00 & 48.23 & 0.2939 & 0.4028 & 0.5784 & 0.2386 \\
& EAGLE-2~\cite{li2024eagle2} & 0.92$\times$ & 1.22 & 47.80 & 0.2937 & 0.4040 & 0.5594 & 0.2385 \\
& LANTERN ($\delta=0.2$)~\cite{jang2024lantern} & 1.26$\times$ & 1.83 & 50.23 & 0.2910 & 0.4458 & 0.5360 & 0.2314\\
& VVS-U ($i=2, \delta=0.1$) & \textbf{1.76$\times$} & \textbf{2.86} & \textbf{47.19} & 0.2886 & \textbf{0.4856} & 0.4588 & 0.2172 \\

\bottomrule
\end{tabular}

}
\end{table*}

\begin{table}[htbp]
\centering
\caption{Impact of token path selection and truncation strategies on quality and speedup.
Verification skipping strategy: uniform skipping with interval $=3$; Conf.:selecting token path with the highest confidence score; Samp.: sampling token path uniformly. Trunc.: whether the selected token path is truncated.}
\label{tab:abla_token}
\setlength{\tabcolsep}{4pt}
\Large
\resizebox{235pt}{!}
{
\begin{tabular}{lcccccc}
\toprule
% \midrule

\textbf{Strategy} & \textbf{Trunc.} & \textbf{TPF} & \textbf{FID $\downarrow$} & \textbf{CLIP score $\uparrow$} & \textbf{Precision $\uparrow$} & \textbf{Recall $\uparrow$} \\
\midrule
Samp. ({$\delta=0.1$})   & \ding{51} & 2.16 & \cellcolor{green!20}\textbf{23.88} & 0.3208 & 0.5498 & 0.6734 \\
Conf. ({$\delta=0.1$})   & \ding{55} & 2.42 & \cellcolor{red!20}26.24 & 0.3201 & 0.5492 & 0.6300 \\
Conf. ({$\delta=0.1$})   & \ding{51} & 2.16 & 24.71 & 0.3206 & 0.5426 & 0.6604 \\
\midrule
Samp. ({$\delta=0.2$})   & \ding{51} & 2.43 & \cellcolor{green!20}\textbf{25.49}  & 0.3210 & 0.5376 & 0.6184 \\
Conf. ({$\delta=0.2$})   & \ding{55} & 2.69 & \cellcolor{red!20}27.41   & 0.3196 & 0.5606 & 0.5988 \\
Conf. ({$\delta=0.2$})   & \ding{51} & 2.43 & 26.20   & 0.3213 & 0.5458 & 0.6110 \\

% \midrule
\bottomrule
\end{tabular}
}
\end{table}

\begin{table}[htbp]
\centering
\caption{Impact of stale feature reuse strategies on generation efficiency and quality. Verification skipping strategy: uniform skipping with interval $=3$; Relaxation threshold: $\delta=0.2$.}
\label{tab:ablation_stale}
\setlength{\tabcolsep}{4pt}
\Large
\resizebox{235pt}{!}
{
\begin{tabular}{lcccccc}
\toprule
% \midrule

$\boldsymbol{s_1}$ & $\boldsymbol{s_2}$ & \textbf{TPF} & \textbf{FID} ($\downarrow$) & \textbf{CLIP Score} ($\uparrow$) & \textbf{Precision} ($\uparrow$) & \textbf{Recall} ($\uparrow$)\\
\midrule
0 & 0 & 2.23 & 32.63 & 0.3113 & 0.5296 & 0.5682 \\
-1 & 3 & 2.27 & 28.93 & 0.3152 & 0.5290 & \textbf{0.6360} \\
-1 & 0 & \textbf{2.31} & \textbf{27.69} & \textbf{0.3179} & \textbf{0.5398} & 0.6088 \\

% \midrule
\bottomrule
\end{tabular}
}
\end{table}

\paragraph{Feature Staleness}

When skipping verification leads to missing features, we retrieve the most recent features from the cached features according to the number of verification-free tokens. 
We compared three strategies to show the effectiveness of our choice: (i) using the most recent stale features ($s_1=s_2=0$), (ii) blending fresh features ($s_1=-1$) and even staler features ($s_2=3$), and (iii) blending fresh ($s_1=-1$) and the most recent cached features ($s_2=0$).
The results in \cref{tab:ablation_stale} demonstrate that our strategy of blending the most recently cached features with fresh ones yields the highest TPF of 2.31 during SD generation while better preserving image quality.

\section{Related Work}
\label{sec:related_work}

Speculative decoding, a lossless acceleration technique for AR models, stands orthogonal to general speedup methods such as quantization~\cite{chen2022mix} and token pruning~\cite{prance2025,li2025spvla}. 
While it has shown remarkable success in the language domain~\cite{cai2024medusa,li2024eagle1,xia2024unlocking,xia2023speculative,sun2023spectr}, its rigid verification prevents even a highly capable draft model from aligning well with the target model~\cite{bachmann2025judge}. 
% Judge Decoding~\cite{bachmann2025judge} introduces an additional judge layer to decide whether draft tokens should be accepted, thereby relaxing the strict constraints from target model. 
In AR image generation, the distributional discrepancy between the draft and target models becomes even more pronounced. 
With the standard verify scheme, SD often fails to deliver any speedup. 
Expanding the acceptance set by leveraging visual token interchangeability can effectively improve the acceleration.
SJD \cite{teng2024SJD} enhances traditional Jacobi Decoding by introducing a probabilistic token acceptance criterion specifically designed for visual AR generation.
LANTERN \cite{jang2024lantern} employs relaxed acceptance, aggregating probabilities of spatially adjacent tokens to improve candidate acceptance rates. 
GSD \cite{so2025GSD} boosts acceptance rates of SJD~\cite{teng2024SJD} through token clustering during verification.
However, these approaches primarily adapt the traditional verify-then-accept paradigm from LLMs to visual AR models. 
In contrast, our work proposes a novel partial verification-free SD framework tailored specifically for visual AR models, moving beyond the constraints of the conventional SD paradigm derived from LLMs.
\section{Conclusion}
\label{sec:conclusion}

In this work, we introduce VVS, a novel SD framework that supports partial verification skipping for visual AR models. Since the traditional SD paradigm of “draft one step, then verify one step” does not explicitly reduce the target model forward passes—the dominant source of inference latency, for the first time, we explore the feasibility of a partial verification-free SD paradigm for accelerating visual AR generation. 
We identify two key observations: \textbf{(i)} {verification redundancy} and \textbf{(ii)} {stale feature reusability}, which make VVS possible for accelerating visual AR models. 
To maintain generation quality while accelerating generation, we design (1) a verification-free token selector with dynamic truncation, (2) token-level feature caching and reuse to enable the candidate construction, and (3) a fine-grained scheduling policy for verification-free steps to trade off quality and speed better. 
Experiments demonstrate that our method cuts the number of target model forward passes by up to $\bm{2.8\times}$ relative to vanilla AR decoding while fully preserving visual fidelity, providing a complementary acceleration route to relaxation-based strategies.
Since our design requires no modification on the standard draft model training strategy~\cite{li2024eagle2}, future research could explore specialized optimizations for draft model training to further accelerate SD with the partial verification skipping mechanism.

\section*{Acknowledgment}

This work is supported in part by National Natural Science Foundation of China(Grant No.~62472249 and 92467204), Shenzhen Science and Technology Program (Grant No.~JCYJ20220818101014030 and~KJZD20240903102300001). 
We thank the anonymous reviewers for their efforts, which have helped improve the quality of this paper.

{
    \small
    \bibliographystyle{ieeenat_fullname}
    \bibliography{main}
}

% WARNING: do not forget to delete the supplementary pages from your submission 
\clearpage
\setcounter{page}{1}
\maketitlesupplementary
\label{appendix}

\setcounter{section}{0}
\section{Implement Details}
\label{app:implement}

We present the pseudocode of the VVS framework in~\cref{alg:skip_speculative} to further illustrate our design.

\renewcommand{\algorithmiccomment}[1]{\hfill{\(\triangleright\)~#1}}
\begin{figure}[htbp]
\begin{algorithm}[H]
\small
\begin{algorithmic}[1]
\Require $\wp$: text prompt;
         $\mathcal{M}_T$: target model;
         $\mathcal{M}_D$: drafter model;
         $L$: max generated length;
         $V_{\text{last}}$: whether last step was verified
\Ensure Generated token sequence $\mathcal{S}$ for decoding to image

\State Initialize feature cache $\mathcal{C}$ and prefill $\wp$ into $\mathcal{M}_T$
\State $\mathbf{a} \leftarrow \mathcal{M}_D(\wp)$        \Comment{initial accepted candidate}
\State $V_{\text{last}} \leftarrow \text{False}$

\For{$t = 0 \dots L-1$}
    \State $f_t \leftarrow \text{Retrieve}(\mathcal{C}, \mathbf{a})$        \Comment{retrieve features for drafting}
    \State $C_t \leftarrow \mathcal{M}_D(f_t, \mathbf{a})$      \Comment{draft candidates}
    \State $\mathbf{skip} \leftarrow \text{Decision}(C_t, t, V_{\text{last}})$  \Comment{\cref{alg:dynamic_skip}}

    \If{$\neg\ \mathbf{skip}$}
        \State $\mathbf{logits}_t, \mathbf{h}_t \leftarrow \mathcal{M}_T(C_t)$
        \State $\mathbf{a} \leftarrow \text{Verify}(\mathbf{logits}_t, C_t)$   \Comment{verification}
        \State $\mathcal{C} \leftarrow \text{UpdateCache}(\mathbf{h}_t)$   \Comment{cache new features}
        \State $V_{\text{last}} \leftarrow \text{False}$
    \Else
        \State $\mathbf{a} \leftarrow \text{UniformSample}(C_t)$   \Comment{verfication-free}
        \State $V_{\text{last}} \leftarrow \text{True}$
    \EndIf
    \State $\mathcal{S} \leftarrow \mathcal{S} \circ \mathbf{a}$   \Comment{concatenate output}
\EndFor
\State \Return $\mathcal{S}$
\caption{VVS with Partial Verification Skipping}
\label{alg:skip_speculative}
\end{algorithmic}
\end{algorithm}
\end{figure}

\section{Supplementary Results of Drafting Stage Analysis}
\label{app:observes}

In \cref{tab:observe_token}, the results demonstrate that after substituting the verification outcomes, both the acceleration performance and generation quality of SD remain highly stable. \cref{tab:observe_feature_reuse} and \cref{tab:observe_feature_blend} further illustrate the impact of various staleness features, highlighting their reusability.

% Please add the following required packages to your document preamble:
% \usepackage{multirow}
\begin{table}[ht]
\centering

\caption{Verification redundancy experiment. $r$ represents the proportion of verified results that are replaced for all iterations. We replace the verified tokens of the target model with the same number of tokens from the token path with the highest confidence score.}
\label{tab:observe_token}

\Large
\resizebox{235pt}{!}{
\begin{tabular}{cccccc}

\toprule
\midrule

$\boldsymbol{r}$ & TPF & \textbf{FID} ($\downarrow$ ) &  \textbf{CLIP Score} ($\uparrow$) & \textbf{Precision} ($\uparrow$) & \textbf{Recall} ($\uparrow$)\\ 
\midrule

% \multirow{5}{*}{\parbox[t]{3cm}{\centering LlamaGen-XL\\Stage I}}
0\%   & \cellcolor{yellow!20}2.27 & 26.63 & 0.3223 & 0.5370 & 0.5962 \\
25\% & \cellcolor{yellow!20}2.27 & 26.36 & 0.3218 & 0.5344 & 0.6118 \\
33\% & \cellcolor{yellow!20}2.27 & 26.23 & 0.3225 & 0.5334 & 0.6120 \\
50\% & \cellcolor{yellow!20}2.27 & 26.55 & 0.3225 & 0.5364 & 0.6128 \\
100\% & \cellcolor{yellow!20}2.25 & 26.19 & 0.3220 & 0.5502 & 0.6158 \\
 
\midrule
\bottomrule
 
\end{tabular}
}

\end{table}

% Please add the following required packages to your document preamble:
% \usepackage{multirow}
\begin{table}[t]
\centering
\caption{Feature reusability experiment results. $s$ denotes the extra staleness introduced for cached features; $s=-1$ indicates using fresh features for each step; $s=0$ indicates using the most recent features cached from prior steps; $s=+i$ indicates using features with additional staleness $i$ compared with $s=0$.}
\label{tab:observe_feature_reuse}

\Large
\resizebox{235pt}{!}{
\begin{tabular}{cccccc}

\toprule
\midrule

$\boldsymbol{s}$ & $\boldsymbol{\tau}$ & \textbf{FID} ($\downarrow$) & \textbf{CLIP Score} ($\uparrow$) & \textbf{Precision} ($\uparrow$) & \textbf{Recall} ($\uparrow$)\\

\midrule

% \multirow{6}{*}{LlamaGen-XL Stage I} 
-1 & 2.27 & 26.53 & 0.3217 & 0.5328 & 0.6126 \\
0  & \cellcolor{yellow!20}1.66 & 24.13 & 0.3215 & 0.5354 & 0.6685 \\
+1 & 1.57 & 23.80 & 0.3210 & 0.5352 & 0.6990 \\
+2 & 1.52 & 23.96 & 0.3210 & 0.5260 & 0.6850 \\
+3 & 1.49 & 23.89 & 0.3206 & 0.5324 & 0.6950 \\ 

\midrule
\bottomrule
 
\end{tabular}
}

\end{table}

\begin{table}[t]
\centering
\caption{Feature blending experiment results. $s$ denotes the extra staleness for cached features; $s_1$: staleness of type $1$ features, $s_2$: staleness of type 2 features; $s=-1$ indicates using fresh features for each step; $s=0$ indicates using the most recent features cached from prior steps; MAL: mean accept length.}
\label{tab:observe_feature_blend}

\Large
\resizebox{235pt}{!}{
\begin{tabular}{ccccccc}

\toprule
\midrule

$\boldsymbol{s_1}$ & $\boldsymbol{s_2}$ & MAL & \textbf{FID} ($\downarrow$) & \textbf{CLIP Score} ($\uparrow$) & \textbf{Precision} ($\uparrow$) & \textbf{Recall} ($\uparrow$)\\
\midrule

-1 & -1 & 2.27 & 26.53 & 0.3217 & 0.5328 & 0.6126 \\
-1 & 0 & \cellcolor{yellow!20}1.93 & 25.17 & 0.3223 & 0.5402 & 0.6516 \\
-1 & 1 & \cellcolor{yellow!20}1.89 & 25.14 & 0.3218 & 0.5336 & 0.6588 \\
-1 & 5 & \cellcolor{yellow!20}1.86 & 24.65 & 0.3215 & 0.5220 & 0.6722 \\
0 & 0 & 1.66 & 24.13 & 0.3215 & 0.5354 & 0.6685 \\

\midrule
\bottomrule
 
\end{tabular}
}
\end{table}

\section{Prompts used in Qualitative Experiment}
\label{prompt}

\begin{figure*}[th]
\centering
\includegraphics[width=0.99\textwidth]{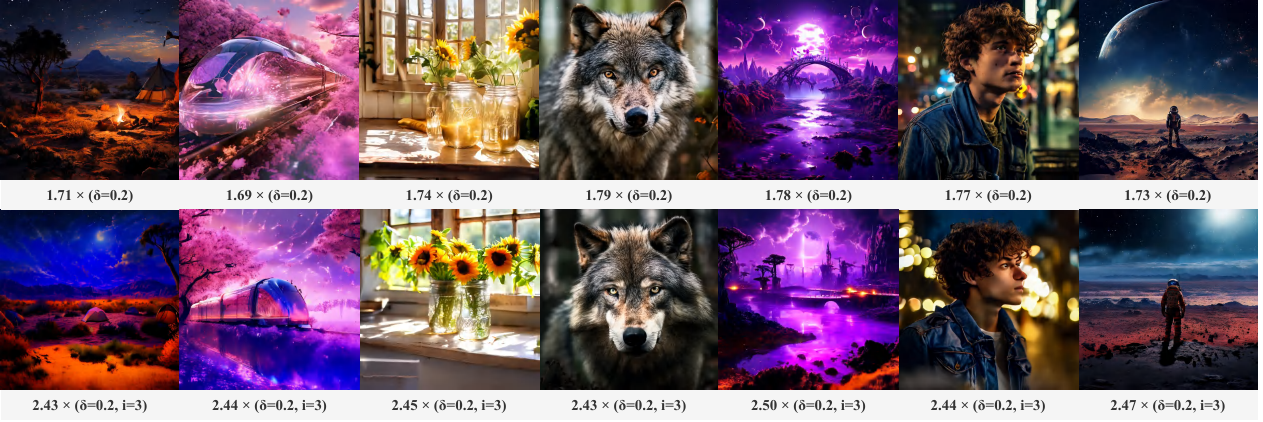}
\caption{Qualitative comparison of results between the acceptance-relax-based SD framework (upper) and ours (lower) on the Lumina-mGPT model. $\delta$ denotes the probability threshold for relaxed acceptance, while $i$ is the interval between two verification-free steps.}
\label{fig:example_figures_lumina}
\end{figure*}

\begin{figure}[h]
  \centering
  \begin{subfigure}[b]{0.52\linewidth}
    \centering
    \includegraphics[width=\linewidth]{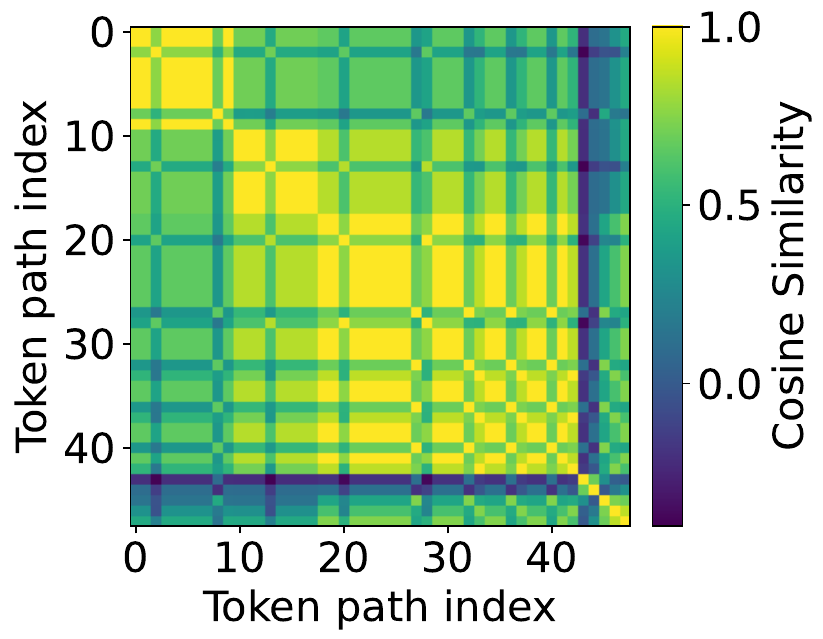}
    \caption{}
  \end{subfigure}%
  \hfill
  \begin{subfigure}[b]{0.42\linewidth}
    \centering
    \includegraphics[width=\linewidth]{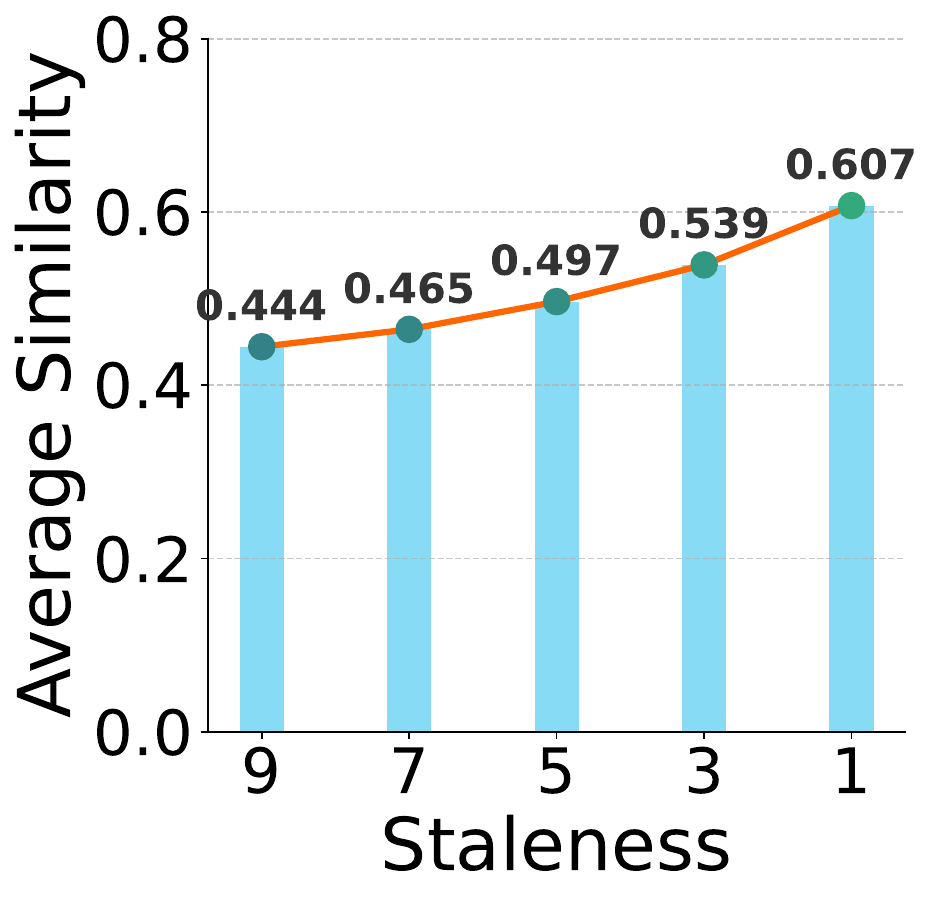}
    \caption{}
  \end{subfigure}
  \caption{For the Lumina-mGPT model: (a) Token path similarity within a candidate token tree. (b) Token-level feature similarity during the SD process. (c) Qualitative results and speedup. }
  \label{fig:token_path_sim_lumina}
  \vspace{-6mm}
\end{figure}

\begin{itemize}
    \item A vast desert landscape under a starry sky, with a single tent illuminated by a warm campfire.
    \item A futuristic glass train speeding through a cherry-blossom forest at sunset, petals swirling in its slipstream.
    \item A rustic farmhouse kitchen with fresh sunflowers in a mason jar, warm afternoon light streaming through open windows.
    \item A close-up of a wolf with intense, focused eyes and thick gray fur, staring directly at the camera, set against a blurred forest background.
    \item A surreal landscape with giant, floating islands connected by glowing energy bridges, all under a purple sky with two moons.
    \item A young man with short curly hair wearing a denim jacket, looking thoughtfully into the distance under city lights.
    \item An astronaut standing on the desolate surface of Mars, gazing at the small, distant blue Earth in the black, star-filled sky.
\end{itemize}

\section{Additional Experiments on Generalization}

We further validated our VVS framework on the Lumina-mGPT model.
\cref{fig:example_figures_lumina} offers a visual demonstration of the resulting image quality, using the same prompts as in \cref{prompt}.
The token path similarity is also relatively high and feature similarity also exists during the SD process, as shown in \ref{fig:token_path_sim_lumina}.
Under the same relaxation threshold $\delta=0.2$, VVS markedly cuts the target model’s forward passes while preserving generation fidelity.
This confirms the broad applicability of our approach across models.
Due to the interchangeability of visual tokens, we believe VVS can be leveraged to accelerate more visual AR models.

\section{System Overhead}

\paragraph{Computation Overhead:} After applying the down-sampling strategy (\S~4.4 \& Alg.~1) and \texttt{torch.jit} optimization, the similarity $s$ computation accounts for $<10\%$ of AR decoding for Lumina-mGPT-7B (6s vs. 70s on a single RTX4090), and this overhead can be fully covered by the latency reduction of VVS.
The depth of drafted token tree is constant and independent of the scale of target model, this overhead ratio diminishes as the model grows, demonstrating excellent scalability.
Moreover, VVS-U supports rule-based scheduling with negligible overhead ($<1\%$).

\paragraph{Memory Overhead:} the constraint on consecutive skips allows VVS maintain only minimal cached features ($n = \max(L_i) = 5$, which is a constant). 
For LlamaGen, with the latest 5 features of 1280 dimensions stored in BF16, the cache footprint is under 100 kB, which is negligible on modern GPUs sporting 24 GB or more of memory.

\section{Discussion of Quantitative Results}
Employing the relaxed acceptance of SD for image AR generation introduces a lossy approximation. 
However, this trade-off is somehow favorable due to:
\textbf{i)} stochastic sampling nature: visual AR generation typically operates with a high temperature, implying that the model inherently explores a diverse distribution rather than a deterministic path.
\textbf{ii)}  visual token interchangeability: 
As noted in GSD and LANTERN, 
multiple diverse tokens can convey equally valid visual semantics.
Therefore, while the relaxation introduces uncertainty $\delta$ (noise), this noise primarily results in switching between semantically similar visual tokens (e.g., slight variations in texture) rather than catastrophic semantic errors.
Thus, SD schemes can even surpass the vanilla AR baseline on certain metrics.
By omitting verification, VVS amplifies noise and, in certain configurations, leads to lower recall and other performance metrics.

\end{document}